\documentclass{egpubl}
\usepackage{pg2026s}

\ConferencePaper        % uncomment for (final) Conference Paper
\WsConferencePaper

\usepackage[T1]{fontenc}
\usepackage{dfadobe}  

\usepackage{cite}  % comment out for biblatex with backend=biber
\BibtexOrBiblatex
\electronicVersion
\PrintedOrElectronic
\usepackage{graphicx}
\ifpdf
\fi

\usepackage{egweblnk} 
\title{GSPotential: Camera Potential Field for Sparse-View 3D Gaussian Splatting}

\author[Z. An et al.]
{\parbox{\textwidth}{\centering
Zeyuan An$^{1}$\orcid{0009-0006-8801-0159}, Yanghang Xiao$^{1}$\orcid{0009-0006-9489-8510}, Zhiying Leng$^{1}$\thanks{Corresponding author.}\orcid{0000-0002-8773-6939}, Yijun Feng$^{2}$\orcid{0000-0002-7026-2036}, and Xiaohui Liang$^{1,3}$\orcid{0000-0001-6351-2538}\\
{\small
$^{1}$State Key Laboratory of Virtual Reality Technology and Systems, Beihang University, Beijing, China\\
$^{2}$School of Computer Science and Engineering, Beihang University, Beijing, China\\
$^{3}$Zhongguancun Laboratory, Beijing, China\\
\texttt{947066339@buaa.edu.cn; sy2406428@buaa.edu.cn; zhiyingleng@buaa.edu.cn}\\
\texttt{buaa\_fengyijun@buaa.edu.cn; liang\_xiaohui@buaa.edu.cn}
}
}}

\usepackage{graphicx}
\usepackage{booktabs}
\usepackage[accsupp]{axessibility}  
\usepackage{multirow}
\usepackage{xcolor}

\begin{document}

% uncomment for using teaser
% \teaser{
%  \includegraphics[width=0.9\linewidth]{images/main.pdf}
%  \centering
%   \caption{New EG Logo}
% \label{fig:teaser}
% }

\maketitle
%-------------------------------------------------------------------------
\begin{abstract}
   3D Gaussian Splatting has achieved remarkable success in photorealistic rendering, yet it suffers from severe overfitting and geometric artifacts in sparse-view scenarios due to the inherent deficiency of photometric supervision. Recent advances have attempted to regularize optimization by incorporating external priors, such as depth, point clouds, or diffusion models. However, these methods typically overlook the non-uniform distribution of supervision across the viewing space, resulting in limited specificity in prior use and primitive control. In this paper, we propose \textbf{GSPotential}, a framework that quantifies view-space supervision imbalance using a Camera Potential Field. Our key insight is to identify supervision valleys where photometric constraints are most deficient, and use the potential field to guide reconstruction from two complementary aspects. First, we devise a probabilistic spherical sampling strategy that places informative virtual cameras in low-potential regions. Point-cloud renderings from these views then provide targeted geometric guidance. Second, the same field provides a directional coverage cue for conservative Gaussian updates in weakly covered spatial sectors. Extensive experiments demonstrate that GSPotential achieves high reconstruction fidelity while maintaining competitive training efficiency.
%-------------------------------------------------------------------------
%  ACM CCS 1998
%  (see https://www.acm.org/publications/computing-classification-system/1998)
% \begin{classification} % according to https://www.acm.org/publications/computing-classification-system/1998
% \CCScat{Computer Graphics}{I.3.3}{Picture/Image Generation}{Line and curve generation}
% \end{classification}
%-------------------------------------------------------------------------
%  ACM CCS 2012
   % (see https://www.acm.org/publications/class-2012)
%The tool at \url{http://dl.acm.org/ccs.cfm} can be used to generate
% CCS codes.
%Example:


\end{abstract}  
%-------------------------------------------------------------------------
\section{Introduction}
\label{sec:intro}

High-quality scene reconstruction from sparse viewpoints is a long-standing challenge in computer vision. While 3D Gaussian Splatting (3DGS) \cite{kerbl20233d} has set new benchmarks for photorealistic rendering, its performance remains heavily dependent on dense multi-view supervision. In sparse-view settings, 3DGS often suffers from severe overfitting and geometric instability, leading to significant artifacts in novel views.

We observe that even when 3DGS is initialized with strong geometric priors such as dense point clouds, rendering can exhibit strong directional instability. 3DGS may fail to preserve the provided geometry during optimization, producing renderings that can even underperform direct projection of the raw point cloud. This suggests that geometric priors are not fully utilized: geometry alone does not guarantee reliable reconstruction without proper guidance across the viewing space.

\begin{figure}[h]
    \centering
    \includegraphics[width=\linewidth]{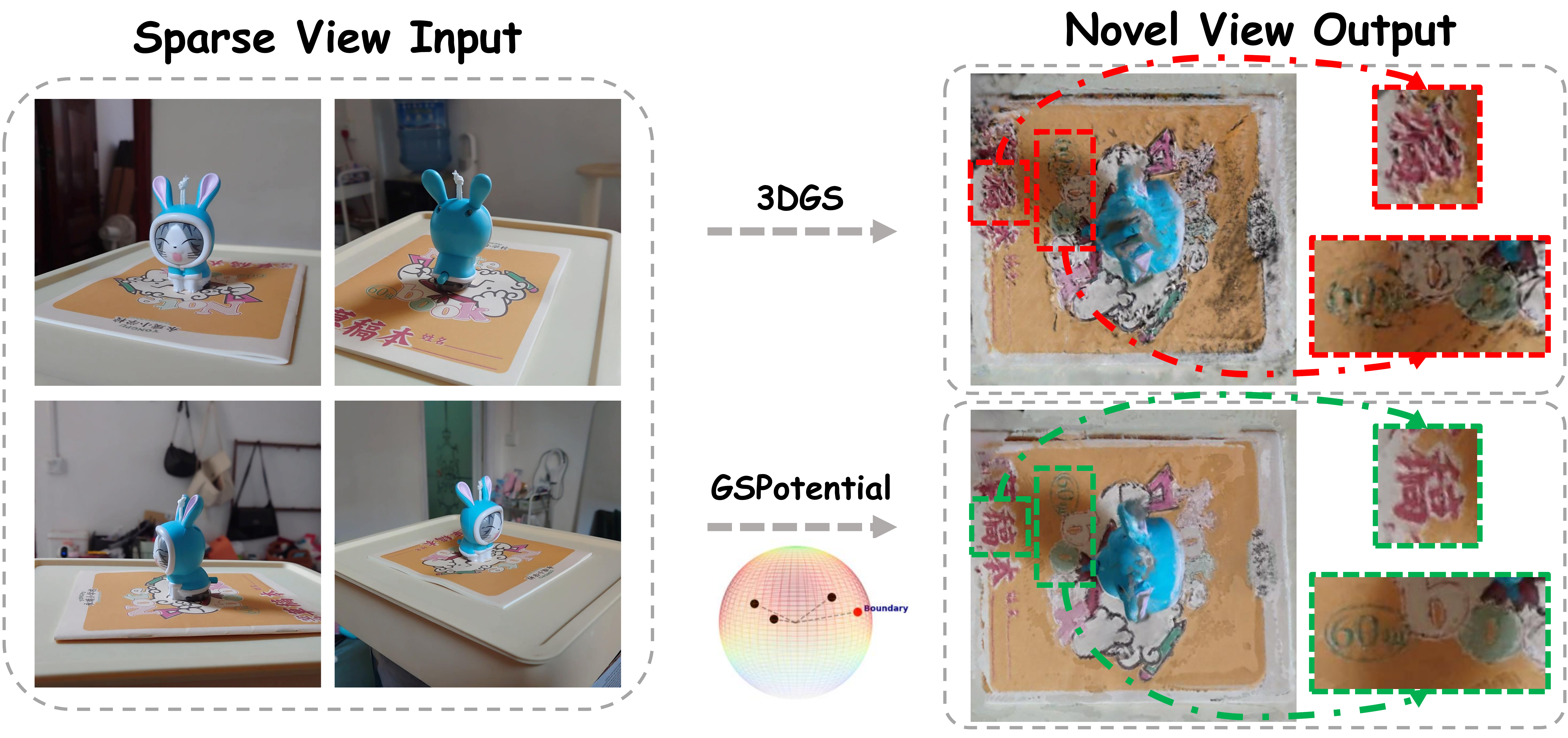}\\
    \caption{Visualizations of scene reconstruction (inputs are from GaussianObject \cite{yang2024gaussianobject}). Given sparse-view inputs, our GSPotential achieves improved reconstruction quality compared to original 3DGS. The boxes highlight regions with visible improvements.}
    \label{fig:comparison}
\end{figure}

Current research primarily addresses sparse-view failure with spatial regularization or generative priors. These methods often treat the viewing space uniformly, applying global constraints without considering the directional distribution of input cameras. Our analysis suggests that geometric instability is closely related to supervision imbalance in view space. In sparse scenarios, input cameras provide highly anisotropic coverage over the viewing sphere. Consequently, optimization is biased toward well-constrained directions, leaving large angular regions under-constrained. This imbalance limits the use of geometric priors in weakly supervised directions.

To address this imbalance, we introduce a Camera Potential Field (CPF) on a canonical viewing sphere. Modeling each input camera as a source whose potential decays with angular distance gives a continuous measure of supervision strength. This identifies weakly supervised yet reliable interpolation regions. We further incorporate a confidence-aware constraint to distinguish them from unobservable blind spots.

Using CPF, we propose \textbf{GSPotential}, which rebalances sparse-view optimization in both supervision and structure. Instead of uniform regularization, our method selectively samples virtual viewpoints from low-potential regions and enforces consistency with point cloud renderings, allowing the geometric prior to act where photometric constraints are most deficient. We also use CPF to derive a directional coverage score for Gaussian primitives, which adjusts pruning and densification in weakly covered sectors. GSPotential thus fills supervision valleys while reducing unsupported primitive growth.

Our contributions are threefold:
\begin{itemize}
\item We identify view-space supervision imbalance as an important factor behind geometric instability in sparse-view 3DGS, and analyze how it limits the effective use of geometric priors.
\item We introduce the Camera Potential Field, a continuous formulation that models supervision strength and confidence over a canonical viewing sphere, enabling the explicit localization of reliable supervision valleys.
\item We propose a potential-guided reconstruction strategy that jointly fills low-potential regions with virtual supervision and regulates Gaussian primitive evolution through coverage-aware pruning and conservative densification.
\end{itemize}

% ----------------

\section{Related Work}

\textbf{Sparse Views 3D Reconstruction.}
Sparse-view reconstruction is a long-standing challenge in computer vision. Traditional Structure-from-Motion (SfM) and Multi-View Stereo (MVS) \cite{seitz2006comparison,schonberger2016structure} methods rely on significant visual overlap, which is often absent in sparse settings. The emergence of 3D Gaussian Splatting \cite{kerbl20233d} has revolutionized real-time rendering, yet it remains prone to overfitting and artifacts when supervision is limited.

Existing sparse-view 3DGS methods target diverse settings. Some focus on small-baseline reconstruction from a limited viewing frustum \cite{zhang2024cor}, while others address sparse $360^{\circ}$ scenes typically using 12--24 images \cite{xiong2024sparsegs,paliwal2024coherentgs}. For extreme few-shot settings, feed-forward approaches reconstruct scenes from as few as two images \cite{charatan2024pixelsplat,szymanowicz2024splatter}. Recent feed-forward or pose-free methods further exploit learned priors or cross-view matching cues \cite{freesplatter,spf_splat}. However, these methods require extensive pre-training, are often tailored to pair-wise or local-view protocols, and may not suit per-scene sparse $360^{\circ}$ optimization.

Our work addresses extreme sparse $360^{\circ}$ reconstruction from only 4 views, where large angular gaps create severe supervision voids. To mitigate performance degradation, previous work have introduced various external priors, including monocular depth estimation \cite{paliwal2024coherentgs,li2024dngaussian,chung2024depth,zhu2023fsgs,zhang2024cor}, dense point cloud initialization \cite{fan2024instantsplat,chen2024dense}, diffusion-based appearance refinement \cite{xiong2024sparsegs,yang2024gaussianobject,gsgs}, or structural regularization on Gaussian primitives \cite{park2025dropgaussian,d2gs}. Surface-oriented sparse-view methods also improve geometry by enforcing depth, feature, or surface consistency \cite{fatesgs,sparse2dgs,sparsesurf}. These priors are typically applied uniformly across the viewing space, without accounting for supervision imbalance. In contrast, GSPotential quantifies this imbalance with CPF and targets geometric priors where they are most needed.
Some methods attempt to bridge the gap by sampling novel viewpoints for regularization, but they often rely on simple linear interpolation between existing poses, stochastic pose perturbation, or generative pseudo views \cite{zhu2023fsgs,xiong2024sparsegs,chung2024depth,yang2024gaussianobject,gsgs}. These approaches emphasize view quantity rather than supervision balance, whereas we explicitly model view-space imbalance to allocate geometric guidance.

\begin{figure*}[t]
\centering
\includegraphics[width=\linewidth]{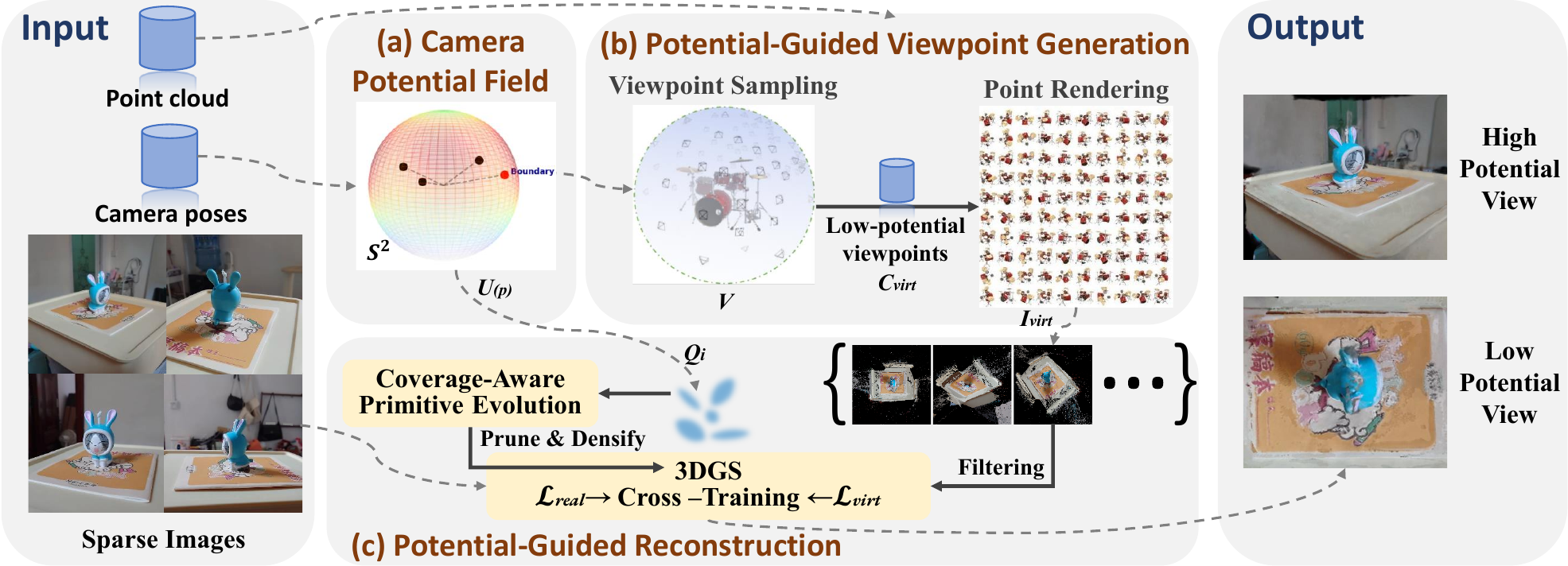}
\caption{Overview of GSPotential. Given sparse images, camera poses, and point clouds, we first estimate a \textbf{Camera Potential Field} to model view-space supervision imbalance. Based on this field, we perform \textbf{Potential-Guided Viewpoint Generation} to generate virtual cameras in low-potential regions. The sampled views are then used for \textbf{Potential-Guided Reconstruction}, while the projected potential provides a directional coverage score for Coverage-Aware Primitive Evolution, including anisotropy pruning and conservative densification.}
\label{fig:overview}
\end{figure*}

\textbf{Geometry Priors and Point-based Supervision.}
Geometric priors, particularly point clouds, are widely used to stabilize neural rendering. Feedforward models such as DUSt3R \cite{wang2024dust3r} and VGGT \cite{wang2025vggt} provide high-quality dense point clouds even from a few uncalibrated images. Many 3DGS methods leverage these point clouds for initialization \cite{fan2024instantsplat,yang2024gaussianobject} or as a global geometric constraint during optimization \cite{chen2024dense}. Recent sparse-view surface reconstruction methods further exploit MVS points, monocular depth, or feature consistency to improve geometric fidelity \cite{sparse2dgs,sparsesurf}.

Most frameworks treat the point cloud as a static or global prior, whose geometry may be under-utilized in directions lacking photometric supervision. Existing geometric constraints also rarely consider how camera distribution affects directional reliability. Our potential field instead invokes geometric guidance in under-constrained directions without over-smoothing well-observed areas. We further use the same field to guide conservative structural updates in weakly covered spatial sectors.

\textbf{View Planning and Camera Modeling.}
The spatial distribution of cameras is a core interest in Active Vision and Next-Best-View (NBV) research \cite{kriegel2012next}. These methods typically use information-theoretic metrics, such as entropy or mutual information, to guide sensor placement for maximizing scene coverage. However, they are designed for data acquisition, whereas sparse-view reconstruction deals with fixed datasets.

We instead model the supervision potential induced by a fixed camera set. The continuous Camera Potential Field links active view planning with passive reconstruction. This directional analysis identifies supervision gaps in view space while providing a camera-layout-aware cue for stabilizing weakly covered regions.

% ------------

\section{Methodology}
\label{sec:Method}
\textbf{Problem Formulation and Overview.}
Given $N$ sparse input images $\{I_i\}_{i=1}^{N}$ and camera poses $\{C_i\}_{i=1}^{N}$, we reconstruct a 3D scene using 3D Gaussian Splatting. In such sparse-view scenarios, the primary challenge arises from the anisotropic coverage of the view space. Unlike dense-view settings where cameras provide near-uniform supervision, a small $N$ often results in clustered camera orientations, leaving vast angular regions under-constrained.

We parameterize viewing directions on a unit sphere $\mathcal{S}^2$ centered at the scene. In typical inward-looking scenarios, each camera $C_i$ can be represented as a unit direction vector $v_i \in \mathcal{S}^2$ pointing from the canonical scene center to the camera position. This spherical formulation decouples supervision analysis from scene scale and complex geometry.

We define the density of these observation vectors on $\mathcal{S}^2$ as the view-space supervision distribution. Our method targets regions lacking sufficient multi-view constraints and introduces the Camera Potential Field to quantify this directional imbalance.

As illustrated in Figure~\ref{fig:overview}, \textbf{GSPotential} estimates a camera potential field and uses it to guide sparse-view reconstruction:
\begin{enumerate}
\item \textbf{Camera Potential Field:} We define a continuous scalar field $U(p)$ over $\mathcal{S}^2$ that integrates supervision strength from training views.
\item \textbf{Potential-Guided Viewpoint Generation:} Based on the inverse of the potential field, we strategically sample virtual viewpoints in low-potential regions that are most prone to reconstruction failure.
\item \textbf{Potential-Guided Reconstruction:} We inject geometric supervision into the sampled low-potential views through point-cloud-based cross-training, and further use the potential field as a camera-layout-aware structural prior to regulate Gaussian pruning and densification.
\end{enumerate}

The potential field thus indicates where auxiliary supervision is needed and where Gaussian updates should be more conservative. The same signal also couples virtual-view allocation with Gaussian primitive evolution. By filling supervision valleys while avoiding unsupported primitive growth, GSPotential stabilizes sparse-view 3DGS optimization.

\subsection{Camera Potential Field}
We model supervision distribution based on the strong correlation between 3DGS degradation and the angular proximity of novel views to training cameras. We therefore project viewing directions onto a continuous scalar field over $\mathcal{S}^2$.

For inward-looking scene scenarios, the complex geometry and varying scales make it difficult to define a uniform supervision metric in 3D space. By adopting a spherical parameterization, we treat each camera as a directional observer. This simplification allows us to analyze the coverage of the scene independently of depth and scene complexity, focusing on the angular density of multi-view constraints.

\textbf{Geometric Canonicalization.} 
Defining the sphere origin is non-trivial because sparse point clouds may be incomplete or contain artifacts that shift their centroid. We therefore estimate a Pseudo-Focus center $c$ from camera extrinsics. Let $o_i$ denote the optical center of camera $i$, and $d_i$ its unit viewing direction. We find $c$ by minimizing the sum of squared distances to all camera optical axes:
\begin{equation}c = \arg\min_{x} \sum_{i} | (x - o_i) \times d_i |^2.\end{equation}
This yields the point closest to the intersection of all camera rays, representing the region of interest most consistently monitored by the input views.

Regarding the sphere radius $R$, while the median distance provides robustness, we adopt the maximum distance among cameras to $c$, $R = \max_i \|o_i - c\|$. This choice ensures that virtual viewpoints remain at a safe distance from the scene. In 3DGS, rendering from a slightly farther distance is significantly more stable than penetrating the scene geometry, which occurs if the radius is too small. Virtual cameras are constrained to this sphere and oriented toward $c$ to maintain a consistent scale and viewing perspective.

\begin{figure}[t]
 \centering 
 \includegraphics[width=\columnwidth]{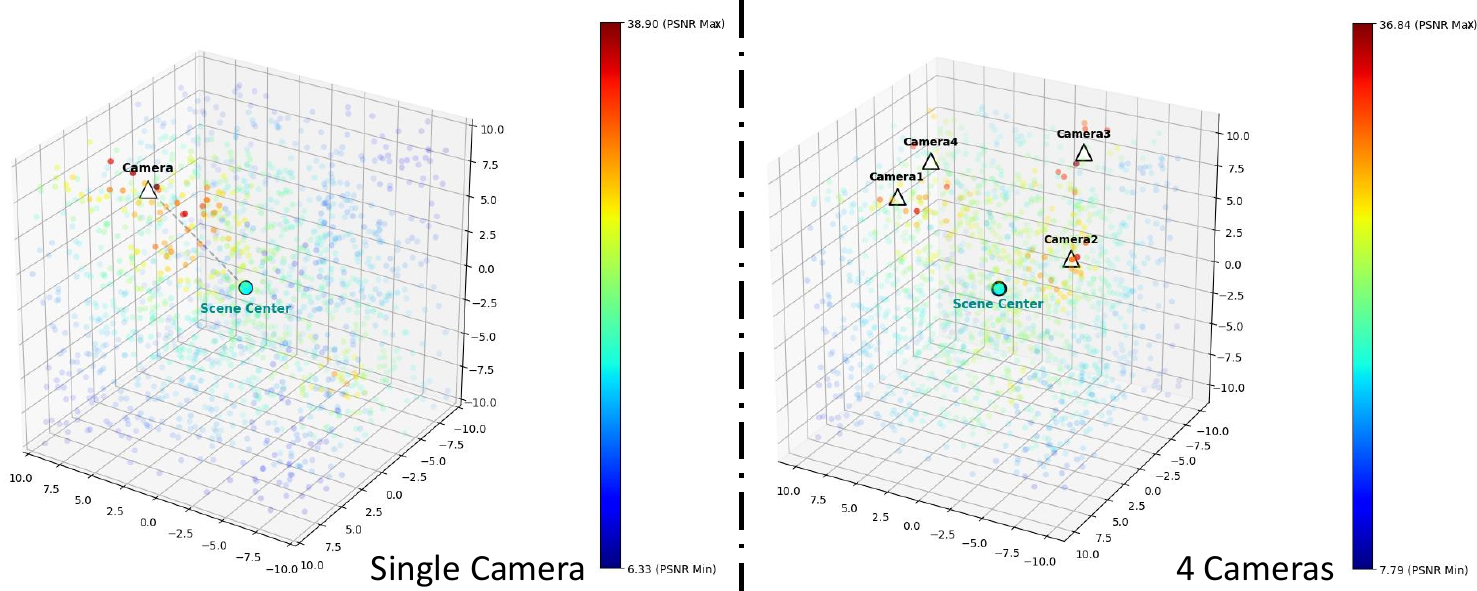}
 \caption{Camera potential fields driven by sampling. We visualize the PSNR of 3DGS rendered images from random viewpoints across view space. 
(Left) In a single-camera setup, high-quality reconstruction is limited to a narrow cone. 
(Right) With 4 sparse cameras, despite the increase in coverage, significant supervision valleys (low PSNR zones) persist across viewpoints.}
 \label{fig:feild1}
\end{figure}

\textbf{Potential Formulation.}
To investigate the directional distribution of reconstruction quality, we conduct a pilot study as illustrated in Fig.~\ref{fig:feild1}. We train a 3DGS model on sparse view sets and densely evaluate the PSNR values across the entire viewing space. 
As shown in Fig.~\ref{fig:feild1}, the resulting fidelity landscape reveals two key phenomena: 
(1) High-fidelity regions are strictly localized around training cameras, forming isolated peaks. 
(2) Outside of the near angular gaps between cameras, the PSNR drops precipitously, forming deep valleys where the optimization fails to maintain structural integrity. 
This correlation between viewpoint proximity and rendering fidelity motivates a Camera Potential Field that decays with angular distance.

\begin{figure*}[t]
\centering
\includegraphics[width=\linewidth]{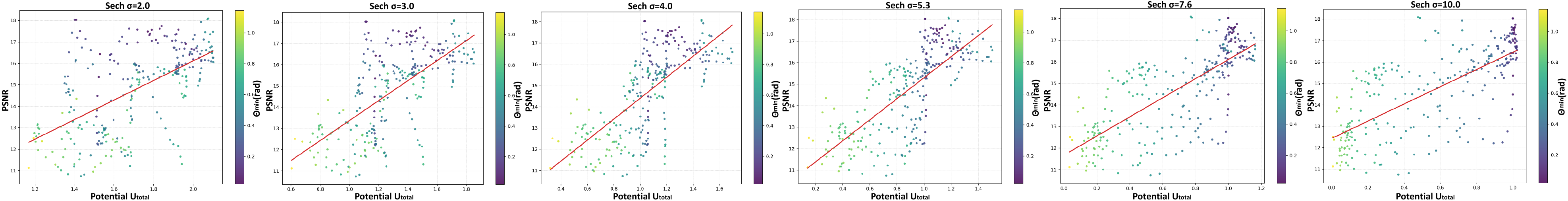}
\caption{Analysis of camera-potential parameterization and novel-view rendering quality. Each scatter point denotes a sampled viewpoint. The horizontal axis is the camera potential computed with a specific parameter setting, and the vertical axis reports the corresponding novel-view rendering quality. Darker colors indicate larger angular distance to the nearest training view. Compared with using angular distance alone, the potential value provides a more discriminative estimate of supervision strength, since viewpoints with similar nearest-view distance can still exhibit different rendering quality under different camera configurations. This verifies that the camera potential captures the joint camera-layout effect beyond a nearest-angle cue.}
\label{fig:com}
\end{figure*}

With canonical center $c$, each training camera maps to $v_i = (o_i - c) / \|o_i - c\|$. For any candidate direction $p \in \mathcal{S}^2$, we quantify its supervision strength by aggregating angular proximity to all training cameras. We define the Information Potential as:
\begin{equation}U_{\text{info}}(p) = \sum_i \mathrm{sech}\left(\sigma \cdot (1 - p^\top v_i)\right).\end{equation}
We determine $\sigma$ from a boundary condition on supervision decay. We require the influence of a single camera to diminish to a baseline margin (1$\%$ of its peak) when the viewing direction reaches the hemisphere boundary ($90^\circ$ from the camera axis). Since $1 - p^\top v_i = 1$ at $90^\circ$, this yields $\sigma = \mathrm{arcsech}(0.01) \approx 5.3$. 
 Fig.~\ref{fig:com} analyzes its relation to rendering quality.

\begin{figure}[t]
 \centering 
\includegraphics[width=\columnwidth]{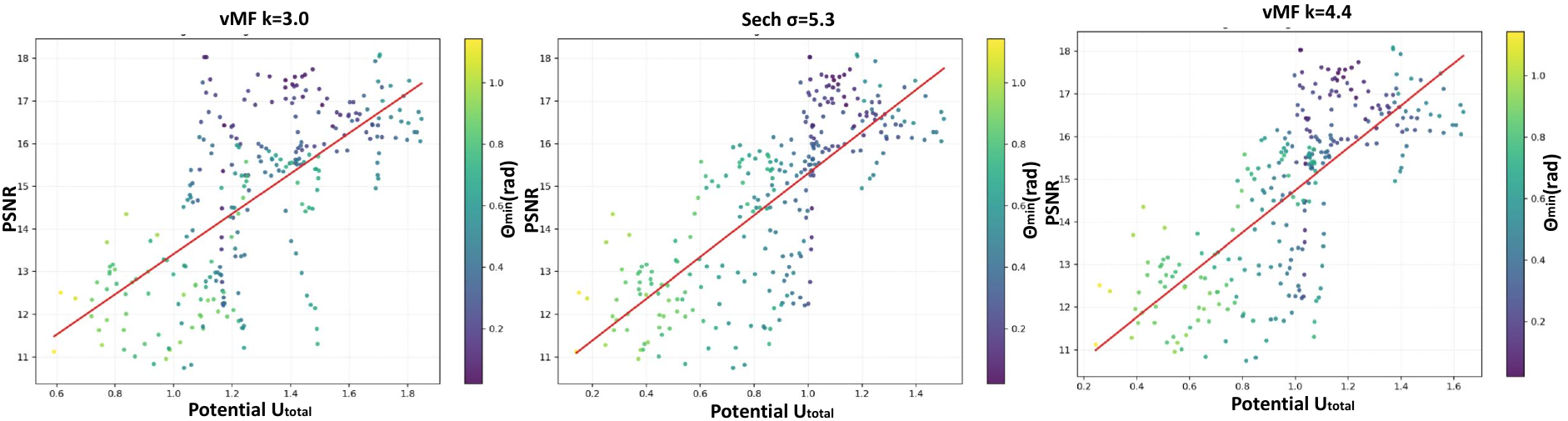}
 \caption{Comparison of kernel choices for modeling camera-potential decay. The horizontal axis $\theta$ denotes the angular distance between a novel viewpoint and its nearest training view. We compare vMF with $\kappa=3.0$ (left), $\mathrm{sech}$ with $\sigma=5.3$ (middle), and vMF with $\kappa=4.4$ (right). The latter two are both calibrated to decay to $1\%$ of their peak at $\theta=90^\circ$, while $\kappa=3.0$ is selected to approximate the overall decay trend of the $\mathrm{sech}$ kernel. Compared with vMF, the $\mathrm{sech}$ kernel yields a smoother relation between camera potential and novel-view rendering quality.}
 \label{fig:vmf_sech}
\end{figure}

On a spherical manifold, a common choice for modeling directional distributions is the von Mises-Fisher (vMF) kernel \cite{gopal2014mises}, which employs exponential decay ($\propto \exp(\kappa p^\top v_i)$). However, when the vMF kernel is calibrated with the same $1\%$ boundary condition at $90^\circ$ (Fig.~\ref{fig:vmf_sech}, right), it decays too rapidly and compresses most weakly observed directions into a narrow low-potential range. This makes the potential less consistent with the gradual degradation of novel-view rendering quality. In contrast, the hyperbolic secant kernel (Fig.~\ref{fig:vmf_sech}, middle) provides a heavier-tailed decay, preserving smoother and longer-range influence across large angular gaps. As a result, the potential values remain more informative in sparse regions and show a more monotonic relationship with rendering quality. Even with $\kappa=3.0$ to better match the $\mathrm{sech}$ decay trend (Fig.~\ref{fig:vmf_sech}, left), vMF remains more scattered, supporting our choice for extreme sparse-view settings.

\textbf{Constraining Extrapolation.} 
Minimizing $U_{\text{info}}(p)$ alone tends to favor directions that are far from all input cameras. However, these blind regions correspond to extrapolation rather than interpolation. In such areas, any prior derived from the same input views (point clouds from DUSt3R or diffusion priors) becomes inherently unreliable due to the lack of observable surface information. To reduce the influence of such directions, we introduce a confidence penalty based on the nearest angular distance to the training views. Let $\theta_{\min}(p) = \min_i \arccos(p^\top v_i)$ be this angular distance, measured in radians. We define:
\begin{equation}U_{\text{conf}}(p) = \theta_{\min}(p),\end{equation}
and the final Camera Potential Field is:
\begin{equation}U_{\text{total}}(p) = U_{\text{info}}(p) + \lambda  U_{\text{conf}}(p).\end{equation}
Low-potential regions with moderate angular support pinpoint directions that are weakly supervised but geometrically plausible. These regions are the primary targets for our subsequent supervision allocation.

\subsection{Potential-Guided Viewpoint Generation}
\label{subsec:sampling}

The potential field $U_{\text{total}}(p)$ provides a continuous measure of supervision deficiency across the viewing sphere. To rectify the imbalanced distribution, we seek to inject auxiliary geometric guidance into regions with insufficient potential.

Selecting only the global minimum of $U_{\text{total}}(p)$ would over-concentrate supervision. Since the supervision gap typically manifests as a continuous angular region rather than an isolated point, such concentration would neglect other weakly-supported directions and potentially cause local overfitting to the geometric prior. To ensure broad spatial coverage and angular diversity, we adopt a probabilistic sampling strategy that favors low-potential regions while exploring the viable interpolation space.

\textbf{Efficient Viewpoint Generation.}
For efficiency, we sample $K$ virtual viewpoints once before training. We first discretize the viewing sphere $\mathcal{S}^2$ into a set of candidate directions $\{p_j\}$ using an approximately uniform grid. For each candidate direction, we assign a sampling weight inversely proportional to its total potential:
\begin{equation}w_j = \frac{1}{U_{\text{total}}(p_j) + \epsilon},\end{equation}
where $\epsilon$ is a small constant for numerical stability. The probability $P_j$ of selecting a direction $p_j$ is then defined as:
\begin{equation}P_j = \frac{w_j}{\sum_m w_m}.\end{equation}
Based on this distribution, we select a fixed set of virtual viewpoints $\mathcal{V} = \{C_{\text{virt}}^{(1)}, \dots, C_{\text{virt}}^{(K)}\}$. This avoids repeated potential evaluation while distributing supervision across low-potential regions.

\subsection{Potential-Guided Reconstruction}
\label{subsec:reconstruction}

We use the potential field to guide reconstruction in two complementary ways. First, geometry-guided cross-training fills low-potential view directions with reliable point-cloud supervision. Second, the projected directional coverage score makes Gaussian updates more conservative in weakly covered spatial sectors. The former compensates for missing supervision, while the latter reduces unsupported primitive growth.

\subsubsection{Geometry-Guided Cross-Training}
The virtual viewpoints $\mathcal{V}$ sampled from low-potential regions identify where the model's self-supervision is weakest. To bridge these gaps, we leverage the geometric prior from the input point cloud $\mathcal{P}$ (generated via DUSt3R \cite{wang2024dust3r}) to provide auxiliary supervision.

\textbf{Pseudo-View Synthesis.}
For each sampled virtual viewpoint $C_{\text{virt}}^{(k)} \in \mathcal{V}$, we render the point cloud $\mathcal{P}$ with point splatting \cite{botsch2003high} to obtain a pseudo image $I_{\text{virt}}^{(k)}$ and a valid mask $M_{\text{virt}}^{(k)}$. The mask records pixels supported by projected points. To handle the sparsity and noise of the point cloud, we apply an Image-Space Connectivity Filter.  We use a $5 \times 5$ neighborhood and remove projected pixels with fewer than two valid neighbors. In this way, virtual supervision is applied only to reliable and spatially coherent point-rendered regions.

The virtual-view loss is simply the standard image loss between the 3DGS rendering and the pseudo image, evaluated only inside the valid mask:
\begin{equation}
\mathcal{L}_{\text{virt}}^{(k)}
=
\mathcal{L}_{\text{img}}(\hat I_{\text{virt}}^{(k)}, I_{\text{virt}}^{(k)} \mid M_{\text{virt}}^{(k)}),
\end{equation}
where $\hat I_{\text{virt}}^{(k)}$ is the 3DGS rendering from $C_{\text{virt}}^{(k)}$. Invalid pixels are ignored in the loss.

\textbf{Confidence-Aware Supervision.}
The reliability of the point cloud prior varies across the view space. As discussed in Sec.~\ref{sec:intro}, while $\mathcal{P}$ is relatively accurate for interpolation, it degrades significantly in extrapolation regions (blind spots). We thus adapt the supervision strength using the confidence term $U_{\text{conf}}(p_k)$ from the potential field. The supervision weight for a virtual view is defined as:
\begin{equation}
\alpha_k = \frac{1}{1 + U_{\text{conf}}(p_k)}.
\end{equation}
This weight gives stronger guidance to virtual views that remain close to the observed camera sectors, while reducing the influence of views that move toward extrapolation regions.

\textbf{Training Objective.}
We alternate between real and virtual views, sampling one of each per iteration and minimizing:
\begin{equation}
\mathcal{L}
=
\mathcal{L}_{\text{real}}
+
\alpha_k \mathcal{L}_{\text{virt}}^{(k)}.
\end{equation}
Here, $\mathcal{L}_{\text{real}}$ is the standard 3DGS loss on the sampled input image, and $\mathcal{L}_{\text{virt}}^{(k)}$ is the masked point-cloud supervision loss from the sampled virtual view. The potential field therefore determines both virtual-view sampling and supervision strength.

\subsubsection{Coverage-Aware Primitive Evolution}
Since the potential field is defined over camera directions, its value on a Gaussian primitive is used as a structural cue rather than a supervision weight. Specifically, we project each Gaussian center onto the canonical sphere and obtain a \emph{directional coverage score}:
\begin{equation}
dr_i = \frac{x_i-c}{\|x_i-c\|}, \qquad
Q_i = U_{\text{info}}(dr_i),
\end{equation}
where $x_i$ is the center of Gaussian $g_i$. The score $Q_i$ indicates whether the spatial sector containing $g_i$ is close to the observed camera sectors. It is not used as a supervision weight; instead, it guides conservative structural updates.

\textbf{Coverage-Aware Anisotropy Pruning.}
Sparse-view 3DGS often creates elongated primitives in weakly constrained regions. Such primitives are not necessarily wrong, since anisotropic Gaussians are useful for representing fine structures, but extreme elongation in weakly covered sectors is more likely to be a floater or needle-like artifact. We measure the anisotropy of Gaussian $g_i$ by
\begin{equation}
a_i =
\frac{\max(s_i)}
{\min(s_i)+\epsilon},
\end{equation}
where $s_i$ denotes its scale along the three principal axes. We then define a coverage-adaptive pruning threshold:
\begin{equation}
\tau_i =
\tau_{\min}
+
(\tau_{\max}-\tau_{\min})
\operatorname{Norm}(Q_i),
\end{equation}
where $\operatorname{Norm}(\cdot)$ normalizes the coverage scores to $[0,1]$ within a scene. A primitive is pruned if
\begin{equation}
g_i \text{ is pruned if } a_i > \tau_i.
\end{equation}
Thus, pruning requires both suspicious anisotropy and weak directional coverage.

\textbf{Conservative Densification.}
We also use $Q_i$ to regulate Gaussian densification. In standard 3DGS, primitives are split or cloned when their accumulated position gradient is large. However, in weakly covered sectors, a large gradient may arise from unreliable sparse-view optimization rather than insufficient model capacity. We therefore raise the densification threshold in low-coverage regions:
\begin{equation}
\delta_i =
\delta_{\min}
+
(\delta_{\max}-\delta_{\min})
\left(1-\operatorname{Norm}(Q_i)\right).
\end{equation}
Let $\nabla_i$ denote the accumulated gradient magnitude used by standard 3DGS densification. We allow densification only when
\begin{equation}
g_i \text{ is densified if } \nabla_i > \delta_i \quad \text{and} \quad a_i \le \tau_i.
\end{equation}
In practice, coverage-aware primitive evolution is enabled after a short warm-up stage, when Gaussian scales become meaningful. This structural policy is complementary to virtual-view supervision: cross-training adds geometric guidance to low-potential view directions, while $Q_i$-guided updates avoid aggressive structural changes in weakly covered spatial sectors.  The resulting thresholds make primitive updates more conservative in weakly covered sectors.

% --------------------

\begin{table*}[t]
\centering
\small
\caption{Quantitative evaluation of our method compared to previous works on two datasets. Results on Mip-NeRF 360 are averaged over seven scenes, and all metrics are computed on uniformly sampled test views. (``--'' denotes unavailable results.)}
\label{tab:Quantitative}
% \resizebox{\linewidth}{!}{
\begin{tabular}{lcccccccc}
\toprule
Dataset & \multicolumn{4}{c}{Mip-NeRF 360} & \multicolumn{4}{c}{OmniObject3D}\\
\cmidrule(lr){2-5} \cmidrule(l){6-9}
\footnotesize Method $|$ Metric & 
\footnotesize PSNR$\uparrow$ & 
\footnotesize SSIM$\uparrow$ & 
\footnotesize LPIPS$\downarrow$ & 
\footnotesize TIME/s$\downarrow$ & 
\footnotesize PSNR$\uparrow$ & 
\footnotesize SSIM$\uparrow$ & 
\footnotesize LPIPS$\downarrow$ & 
\footnotesize TIME/s$\downarrow$ \\
\midrule
3DGS \cite{kerbl20233d} & 13.52 & 0.282 & 0.588 & 500 
     & 20.87 & 0.805 & 0.221 & 314 \\

SparseGS \cite{xiong2024sparsegs} & 11.15 & 0.166 & 0.672 & 627 
         & 19.15 & 0.751 & 0.305 & 745 \\

FSGS \cite{zhu2023fsgs} & 13.49 & 0.324 & 0.573 & 762
     & 20.13 & 0.797 & 0.206 & 575 \\

CoR-GS \cite{zhang2024cor} & 12.90 & 0.309 & 0.574 & 2110
       & 21.44 & 0.813 & 0.197 & 1559 \\

SplatFields \cite{mihajlovic2024splatfields}       
& 11.92 & 0.331 & 0.748 & 2865
& 21.12 & 0.811 & 0.235 & 1194 \\

GaussianObject \cite{yang2024gaussianobject}
               & -- & -- & -- & -- 
               & 20.89 & 0.813 & \textbf{0.180} & 4773 \\

InstantSplat \cite{fan2024instantsplat} & 15.35 & 0.381 & 0.626 & 543 
             & 20.73 & 0.791 & 0.246 & 352 \\

DropGaussian \cite{park2025dropgaussian} & 13.91 & 0.334 & 0.541 & 502 
             & 21.09 & \textbf{0.819} & 0.203 & \textbf{305} \\

SparseSurf \cite{sparsesurf} & 12.71 & 0.239 & 0.648 & 1410 
             & 18.05 & 0.713 & 0.324 & 754 \\

D$^2$GS \cite{d2gs} & 14.32 & 0.372 & 0.534 & \textbf{345} 
             & 20.03 & 0.767 & 0.207 & 402 \\

Ours & \textbf{16.43} & \textbf{0.411} & \textbf{0.505} & 516 
     & \textbf{21.58} & \textbf{0.819} & 0.201 & 325 \\
\bottomrule
\end{tabular}
% }
\end{table*}

\begin{figure}[t]
\centering
\includegraphics[width=\linewidth]{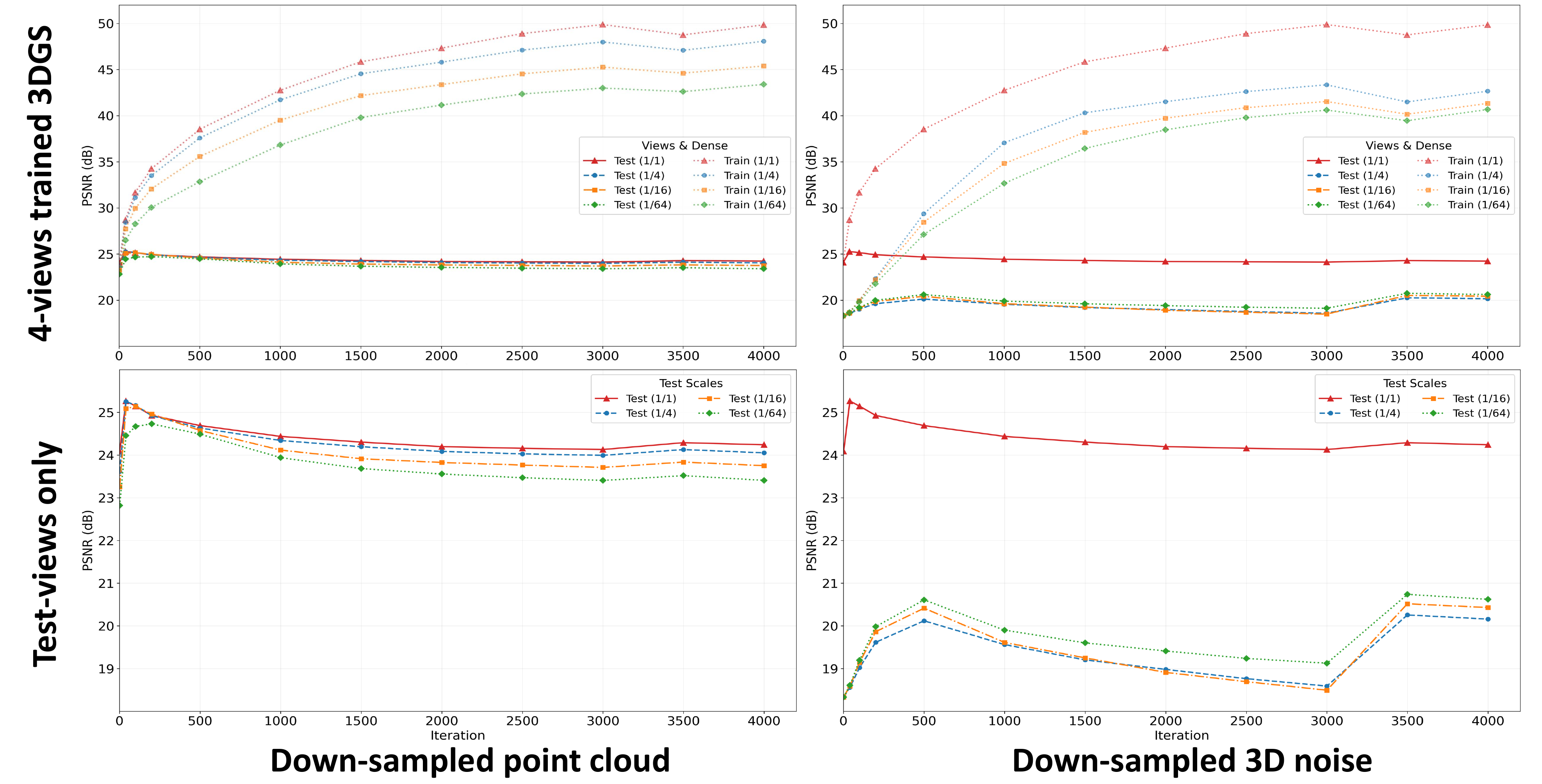}
\caption{
Comparison of 3DGS training with different point cloud initialization strategies. The plots visualize training iterations (x) versus PSNR (y), with: left for original point cloud (generated by DUSt3R), 1/4, 1/16, and 1/64 down-sampling ratios; right for original point cloud (generated by DUSt3R) and random 3D noise initialization with the same down-sampling ratios.
}
\label{fig:veri}
\end{figure}

\section{Experiments}
\label{sec:Experiments}

\textbf{Implementation Details:}
Experiments were conducted on an Nvidia RTX A5000 GPU. We train each model for 10,000 iterations and use the default image resolution setting, with images scaled to a width of 1600 pixels. The penalty scale is set to $\lambda = 0.2$ (selected from our parameter study as it balances view realism and diversity). The number of sampled views is set to $k=256$ (selected as it offers strong coverage of under-constrained viewpoints with reasonable computational cost, while larger values yield negligible improvements).

\textbf{Datasets and Evaluation:}
We evaluate our method on two representative datasets: \textit{Mip-NeRF 360} \cite{barron2022mip} and \textit{OmniObject3D} \cite{wu2023OmniObject3D}. We uniformly select 4 images from each original sequence as the training set and uniformly sample 60 views from the remaining sequence for testing. The reported Mip-NeRF 360 results are averaged over seven scenes. We employ DUSt3R \cite{wang2024dust3r} to estimate camera poses and an initial point cloud from the 4 training images. This "in-the-wild" pipeline aligns with practical capture scenarios where precise poses are unavailable. For a fair comparison, all baselines use the same point cloud, poses, and train--test splits.  All baselines are evaluated under this common 4-view $360^{\circ}$ protocol, including methods originally developed for denser or forward-facing capture.

\textbf{Metrics:} We report \textit{PSNR} \cite{5596999}, \textit{SSIM} \cite{wang2004ssim}, \textit{LPIPS} \cite{zhang2018lpips}, and end-to-end runtime.

\subsection{Verification Study}

\textbf{}

\textbf{Potential and Rendering Quality.}
We analyze the correlation between camera potential and novel-view rendering quality across viewpoints. As illustrated in Figure~\ref{fig:com}, we compare different parameter settings for computing the potential field and plot the resulting potential values against the rendering quality of 3DGS from sampled viewpoints.

The scatter plots show a clear positive relationship between camera potential and rendering quality: viewpoints with higher potential are generally reconstructed more faithfully, while low-potential viewpoints are more likely to suffer from sparse-view artifacts. This supports our key assumption that the potential field captures the directional distribution of photometric supervision.

The color of each point further encodes its angular distance to the nearest training view. Although this distance is an important cue, points with similar nearest-view distance can still have noticeably different rendering quality. This is because supervision strength is affected by the joint configuration of all input cameras, rather than by the closest camera alone. As shown in Figure~\ref{fig:com}, the camera potential separates viewpoints that are mixed under the nearest-angle cue, demonstrating that CPF captures camera-layout-dependent supervision beyond angular distance alone. This observation motivates our use of the potential field for viewpoint generation in Sec.~\ref{subsec:sampling}.

\textbf{Point Cloud Fidelity.}
To investigate how geometric priors mitigate reconstruction ambiguity, we evaluate 3DGS performance under varying initialization conditions. As shown in Figure~\ref{fig:veri}, the density of the initial point cloud directly dictates the reconstruction ceiling; a decimation of points leads to a PSNR drop of up to 6 dB. This confirms that without dense photometric overlap, explicit geometric anchors are the primary source of structural integrity.

Furthermore, we observe that while high-fidelity initializations from DUSt3R provide a significant baseline boost over noise points, vanilla 3DGS fails to fully exploit this geometric prior. As iterations progress, the performance in novel views often plateaus, indicating that the model gradually loses its initial geometric structure in pursuit of overfitting sparse photometric pixels. This observation provides the core motivation for our framework: rather than using the point cloud only for initialization, we must utilize it via potential-guided supervision to prevent the model from deviating into geometrically inconsistent states.

\begin{figure*}[t]
\centering
\includegraphics[width=\linewidth]{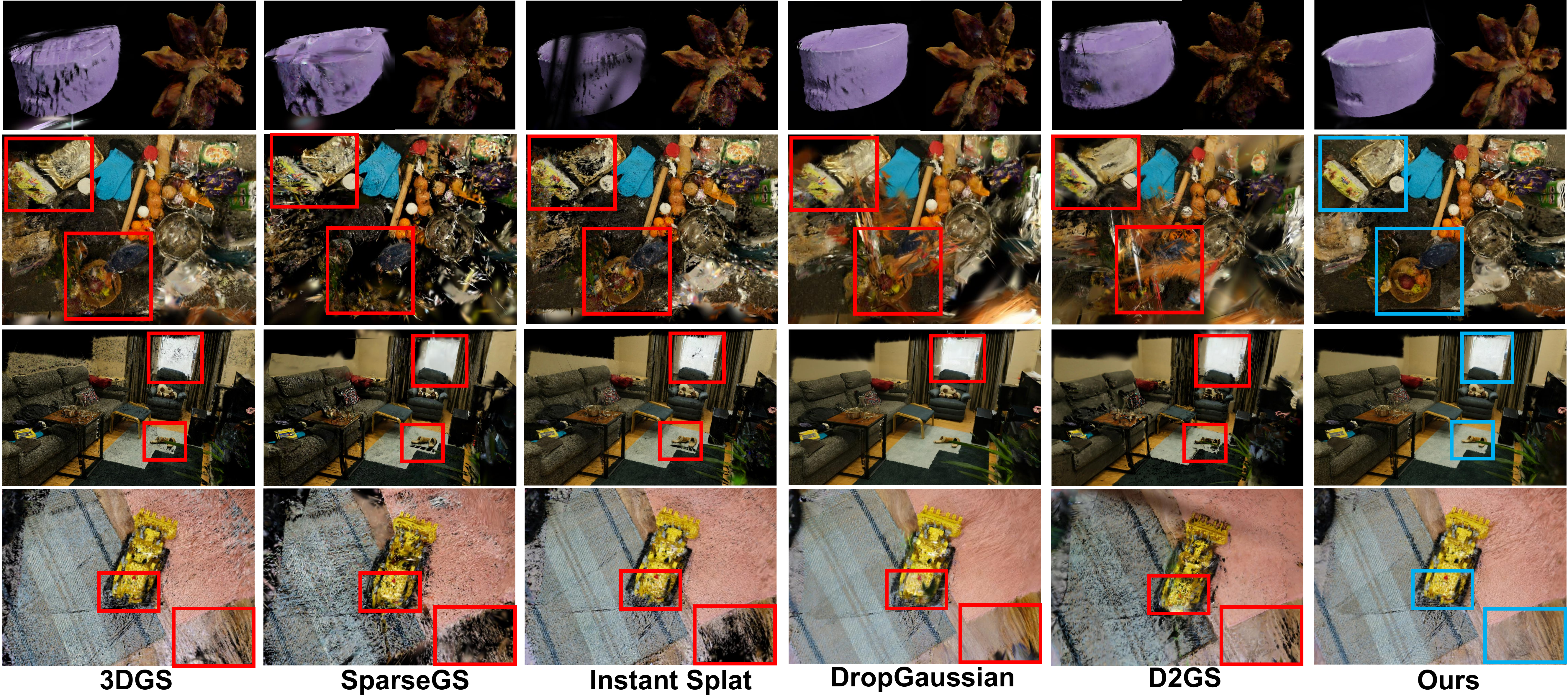}
\caption{Qualitative comparison on unsupervised views. Boxes highlight regions with notable improvements by GSPotential.}
\label{fig:qual}
\end{figure*}

\subsection{Quantitative Comparisons}

Table \ref{tab:Quantitative} compares GSPotential with sparse-view 3DGS methods under the same protocol using \textit{PSNR}, \textit{SSIM}, \textit{LPIPS}, and runtime. For Mip-NeRF 360, we report the average over seven scenes, and all metrics are computed on the uniformly sampled test views.

On Mip-NeRF 360, our method achieves the best performance across PSNR, SSIM, and LPIPS, outperforming both geometry-oriented and structural-regularization baselines.  On OmniObject3D, our method obtains the highest PSNR and matches the best SSIM, while remaining competitive in LPIPS. These results show that GSPotential is effective in both scene-level and object-centric sparse-view reconstruction.

Although our method introduces overhead for camera potential calculation and point cloud rendering, the total time consumption remains close to 3DGS and much faster than several geometry- or surface-oriented baselines. The potential field computation and sampling take approximately 13s, and point cloud rendering takes around 35s for all 256 images. This efficiency is achieved through our image filtering strategy in cross-training, where fewer pixels (only spatially coherent pixels) are computed when using rendered images.

\subsection{Qualitative Comparisons}

As shown in Figure \ref{fig:qual}, our method demonstrates superior visual quality compared to original 3DGS, SparseGS, InstantSplat, DropGaussian, and D$^2$GS. 

On the Mip-NeRF 360 dataset, our approach achieves more accurate scene reconstruction with sharper edges and cleaner planar regions. This is primarily because the potential field directs virtual supervision to fill the gaps between training cameras, preventing the formation of floaters and needle-like artifacts typical of sparse-view 3DGS. For OmniObject3D, our method significantly suppresses drifting and geometric distortion in unsupervised viewpoints, resulting in smoother and more stable surface geometry.

These visual improvements in consistency and stability are direct consequences of our CPF-guided framework. By quantifying the inherent supervision strength across the view space, our method identifies weakly constrained directions and introduces geometric cues where photometric information is insufficient. The resulting reconstructions contain fewer floaters and distorted regions while preserving fine details, maintaining the flexibility of 3DGS together with improved global geometric coherence.

\begin{table}[t]
\centering
\caption{
Ablations of different components of GSPotential, evaluated on Mip-NeRF 360. $U_{\text{info}}$ denotes the information potential used for low-potential viewpoint generation. $Q_i$-P and $Q_i$-D denote coverage-aware pruning and densification guided by the directional coverage score $Q_i$, respectively. $U_{\text{conf}}$ denotes the confidence term for suppressing unreliable virtual supervision near extrapolation regions.
}
\label{tab:ablation}
\resizebox{1.\linewidth}{!}{
\begin{tabular}{ccccc|ccc}
\toprule
$\mathcal{L}_{\text{virt}}$ & $U_{\text{info}}$ & $Q_i$-P & $Q_i$-D & $U_{\text{conf}}$
& PSNR$\uparrow$ & SSIM$\uparrow$ & LPIPS$\downarrow$ \\
\midrule
$\times$ & $\times$ & $\times$ & $\times$ & $\times$ & 13.52 & 0.282 & 0.588 \\
$\checkmark$ & $\times$ & $\times$ & $\times$ & $\times$ & 14.49 & 0.356 & 0.624 \\
$\checkmark$ & $\checkmark$ & $\times$ & $\times$ & $\times$ & 15.90 & 0.379 & 0.601 \\
$\checkmark$ & $\checkmark$ & $\checkmark$ & $\times$ & $\times$ & 16.03 & 0.388 & 0.585 \\
$\checkmark$ & $\checkmark$ & $\times$ & $\checkmark$ & $\times$ & 16.17 & 0.384 & 0.594 \\
$\checkmark$ & $\checkmark$ & $\checkmark$ & $\checkmark$ & $\times$ & 16.34 & 0.394 & 0.577 \\
$\checkmark$ & $\checkmark$ & $\checkmark$ & $\checkmark$ & $\checkmark$ & 16.43 & 0.411 & 0.505 \\
\bottomrule
\end{tabular}
}
\end{table}

\subsection{Ablation Study}

\textbf{Ablation of Camera Potential Field.}
\textit{Effectiveness of Virtual Supervision ($\mathcal{L}_{\text{virt}}$).}
The baseline (Row 1) represents the original 3DGS initialized with DUSt3R point clouds. As shown in Table~\ref{tab:ablation}, simply introducing virtual view supervision $\mathcal{L}_{\text{virt}}$ with uniform interpolation sampling (Row 2) improves PSNR from 13.52 to 14.49. This confirms that point-cloud-rendered virtual views can provide useful geometric guidance under sparse-view supervision. However, without potential-based viewpoint selection, these virtual views may be placed redundantly near already covered regions or in less informative directions, which limits the gain and slightly degrades LPIPS.

\textit{Impact of Information Potential ($U_{\text{info}}$).}
When the information potential $U_{\text{info}}$ is used to guide viewpoint generation (Row 3), the performance increases substantially to 15.90 dB PSNR. This is the largest improvement among all components, confirming that explicitly locating supervision valleys is the key factor of our method. Compared with uniform sampling, $U_{\text{info}}$ better reflects the view-space distribution of camera supervision and allocates virtual supervision to directions where photometric constraints are most deficient.

\begin{figure*}[t]
\centering
\includegraphics[width=\linewidth]{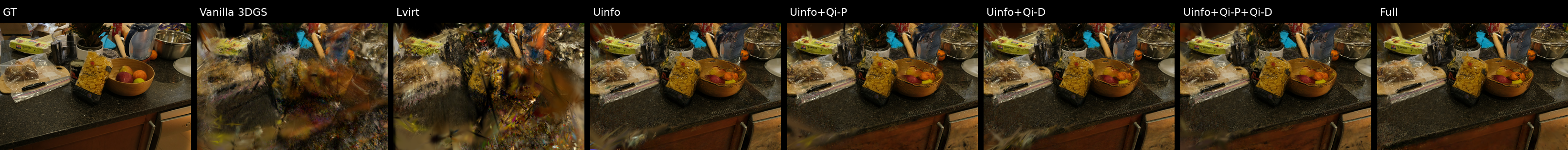}
\caption{Qualitative ablation on test view of the Mip-NeRF 360 \textit{counter} scene. Columns correspond to the variants in Table~\ref{tab:ablation}.}
\label{fig:ab}
\end{figure*}

\textit{Effectiveness of Coverage-Aware Primitive Updates.}
Rows 4--6 evaluate the primitive evolution strategy guided by the directional coverage score $Q_i$. Adding $Q_i$-guided pruning (Row 4) mainly improves SSIM and LPIPS, indicating that removing over-elongated primitives in weakly covered spatial sectors effectively suppresses floaters and needle-like artifacts. In contrast, $Q_i$-guided densification (Row 5) brings a larger PSNR gain, since conservative densification prevents unreliable growth while still allowing sufficient model capacity in well-covered regions. Combining both updates (Row 6) further improves all metrics, showing that pruning and densification play complementary roles in stabilizing Gaussian structure.  Fig.~\ref{fig:ab} shows consistent visual improvements as the proposed components are progressively introduced.

\textit{Effectiveness of Confidence Penalty ($U_{\text{conf}}$).}
Finally, introducing the confidence penalty $U_{\text{conf}}$ (Row 7) further improves the full model, especially in SSIM and LPIPS. This term reduces the influence of virtual supervision near extrapolation regions, where point-cloud renderings are less reliable. As a result, the model avoids fitting misleading pseudo observations while preserving the benefits of potential-guided supervision and coverage-aware primitive evolution.

\begin{table}[t]
\centering
\caption{Ablation of point cloud quality, evaluated on Mip-NeRF 360. Views refer to number of views used to build the point cloud.}
\label{tab:point_performance}
\resizebox{1.\linewidth}{!}{
\begin{tabular}{ccccccc}
\toprule
\multirow{2}{*}{\textbf{Views}} 
& \multicolumn{3}{c}{\textbf{Ours}} 
& \multicolumn{3}{c}{\textbf{3DGS}} \\
\cmidrule(lr){2-4} \cmidrule(lr){5-7} 
 & PSNR $\uparrow$ & SSIM $\uparrow$ & LPIPS $\downarrow$ & PSNR $\uparrow$ & SSIM $\uparrow$ & LPIPS $\downarrow$ \\
\midrule
1        & 15.73 & 0.392 & 0.571 
      & 13.02 & 0.281 & 0.604 \\
2        & 16.14 & 0.408 & 0.527 
      & 13.21 & 0.282 & 0.597 \\
4        & 16.43 & 0.411 & 0.505 
      & 13.52 & 0.282 & 0.588 \\
8        & 16.59 & 0.415 & 0.480 
      & 13.78 & 0.282 & 0.581 \\
\bottomrule
\end{tabular}%
}
\end{table}

\textbf{Ablation of point cloud quality.}
This study evaluates GSPotential under varying point-cloud quality. We generate point clouds using DUSt3R from 1--8 images while training with the original 4 images, and compare against the original 3DGS (as shown in Table \ref{tab:point_performance}). This experiment analyzes sensitivity to the external geometric prior. As expected, increasing the number of images improves point cloud density and coverage, leading to better reconstruction quality for both methods.

Our method consistently outperforms 3DGS across all settings, with larger gains as the point cloud becomes denser and more complete. All main comparisons use point clouds estimated from the same 4 training images. Notably, GSPotential with the weakest point cloud prior still outperforms 3DGS with the strongest one, highlighting its ability to benefit from noisy or incomplete geometry while avoiding the instability observed in direct supervision schemes.  The 8-view point cloud provides only a 0.16 dB gain over the 4-view point cloud, indicating that the DUSt3R prior is already near saturation in this setting.

\textbf{Sampling Strategy and Input-View Sensitivity.}
Table~\ref{tab:sampling_strategies} compares alternative virtual-view sampling strategies on the \textit{kitchen} scene. CPF achieves the best PSNR, SSIM, and LPIPS, showing the benefit of allocating virtual supervision according to the camera distribution. Table~\ref{tab:view_count} further evaluates 3--9 input views. GSPotential consistently outperforms 3DGS across all tested settings, while the absolute performance varies with the camera configuration.

\begin{table}[t]
\centering
\small
\caption{Controlled comparison of virtual-view sampling strategies on the Mip-NeRF 360 \textit{kitchen} scene. All variants use the same reconstruction setting and differ only in virtual-view selection.}
\label{tab:sampling_strategies}
\begin{tabular}{lccc}
\toprule
Sampling & PSNR$\uparrow$ & SSIM$\uparrow$ & LPIPS$\downarrow$ \\
\midrule
Nearest-angle random & 15.63 & 0.318 & 0.520 \\
Uniform sphere & 15.58 & 0.306 & 0.525 \\
Farthest-point sphere & 15.50 & 0.305 & 0.539 \\
Uniform interpolation & 15.52 & 0.310 & 0.522 \\
CPF (ours) & \textbf{15.93} & \textbf{0.373} & \textbf{0.468} \\
\bottomrule
\end{tabular}
\end{table}

% ------------

\section{Limitation and Future Work}

Our current experiments focus on inward-looking and bounded scene reconstruction, where input views approximately observe a shared region of interest.  CPF currently models camera-layout imbalance without explicitly encoding scene geometry, occlusion, or visibility. Extending the formulation to outward-looking scenes and incorporating  geometry-aware visibility cues are natural directions for future work.

Moreover, GSPotential requires an external prior such as a point cloud, typically obtained from feedforward methods such as DUSt3R.  Since CPF depends on camera directions, pose inaccuracies can also perturb the potential field; incorporating pose uncertainty is a possible extension. Exploring more reliable geometry estimation techniques such as VGGT, or refining virtual-view supervision with diffusion models, is also a possible direction for future work.

\begin{table}[t]
\centering
\small
\caption{Sensitivity to the number of input views on the Mip-NeRF 360 \textit{kitchen} scene. Each entry reports PSNR/SSIM/LPIPS.}
\label{tab:view_count}
\begin{tabular}{ccc}
\toprule
Views & 3DGS & GSPotential \\
\midrule
3 & 14.81/0.331/0.508 & 15.49/0.358/0.478 \\
4 & 15.09/0.337/0.495 & 15.93/0.373/0.468 \\
5 & 15.78/0.398/0.460 & 17.00/0.412/0.441 \\
6 & 15.59/0.359/0.495 & 15.89/0.413/0.464 \\
9 & 16.81/0.469/0.441 & 17.19/0.498/0.404 \\
\bottomrule
\end{tabular}
\end{table}

\section{Conclusion}

We presented GSPotential, a framework for addressing view-space supervision imbalance in sparse-view 3D Gaussian Splatting. By introducing the Camera Potential Field, we provide a continuous measure of supervision strength over the viewing sphere, which allows us to identify low-potential regions where photometric constraints are insufficient. Based on this field, our method performs potential-guided viewpoint generation to place virtual cameras in weakly supervised yet geometrically plausible regions, and uses point-cloud renderings to provide targeted geometric guidance.
We further use the same field to derive a directional coverage score for coverage-aware primitive evolution. This score helps regulate anisotropy pruning and conservative densification in weakly covered spatial sectors, improving structural stability and perceptual quality without introducing substantial overhead. Experiments on Mip-NeRF 360 and OmniObject3D demonstrate that GSPotential improves sparse-view reconstruction quality under 4-view settings, and the ablation studies verify the effectiveness of both potential-guided virtual supervision and coverage-aware primitive updates.

\section{Acknowledgements}
This work was supported by the National Natural Science Foundation of China (Project Number: 62272019), supported by the China Postdoctoral Science Foundation under Grant Number 2025M774236, supported by the Postdoctoral Fellowship Program of CPSF under Grant Number GZC20242159 and the fundamental research funds for the central universities.

\clearpage
%-------------------------------------------------------------------------
% bibtex
\bibliographystyle{eg-alpha-doi} 
\bibliography{egbibsample}       

% biblatex with biber
% \printbibliography                

%-------------------------------------------------------------------------
%Color tables are no longer required for purely electronic publications.

\end{document}